# Template Based Inference in Symmetric Relational Markov Random Fields

**Ariel Jaimovich**　　　**Ofer Meshi**　　　**Nir Friedman**
School of Computer Science and Engineering
Hebrew University of Jerusalem
Jerusalem, Israel 91904
{arielj,meshi,nir}@cs.huji.ac.il

## Abstract

*Relational Markov Random Fields* are a general and flexible framework for reasoning about the joint distribution over attributes of a large number of interacting entities. The main computational difficulty in learning such models is inference. Even when dealing with complete data, where one can summarize a large domain by sufficient statistics, learning requires one to compute the expectation of the sufficient statistics given different parameter choices. The typical solution to this problem is to resort to approximate inference procedures, such as loopy belief propagation. Although these procedures are quite efficient, they still require computation that is on the order of the number of interactions (or features) in the model. When learning a large relational model over a complex domain, even such approximations require unrealistic running time.

In this paper we show that for a particular class of relational MRFs, which have inherent symmetry, we can perform the inference needed for learning procedures using a *template-level* belief propagation. This procedure's running time is proportional to the size of the relational model rather than the size of the domain. Moreover, we show that this computational procedure is equivalent to sychronous loopy belief propagation. This enables a dramatic speedup in inference and learning time. We use this procedure to learn relational MRFs for capturing the joint distribution of large protein-protein interaction networks.

## 1 Introduction

Relational probabilistic models are a rich framework for reasoning about structured joint distributions [6, 9]. Such models are used to model many types of domains like the web [22], gene expression measurements [20] and protein-protein interaction networks [11]. In these domains, they can be used for diverse tasks, such as prediction of missing values given some observations [11], classification [22], and model selection [20]. All of these tasks require the ability to perform inference in these models. Since in many models exact inference is infeasible, most studies resort to approximate inference such as variational approximations [12] and sampling [8]. Unfortunately in many cases even these approximations are computationally expensive. This is especially problematic in settings where inference is performed many times, such as parameter estimation.

In this paper we show that we can exploit symmetry properties of relational models to perform efficient approximate inference. Our basic observation is that symmetry in the relational model implies that many of the intermediate results of approximate inference procedures, such as loopy belief propagation, are identical. Thus, instead of recalculating the same terms over and over, we can perform inference at the template level. We define formally a large class of relational models that have these symmetry properties, show how we can use them to perform efficient approximate inference and compare our results with other methods. This is, to the best of our knowledge, the first approximate inference algorithm that works on the template level of the model. However, this efficient inference procedure is limited to cases were we have no evidence on the model, since such evidence can break the symmetry properties. Nevertheless, we show that in many cases, inference with no evidence is useful, especially in learning tasks. Finally, we show a real life application by learning the properties of a model for protein-protein interactions.

## 2 Symmetric relational models

Relational probabilistic models [6, 9, 18, 21] provide a language for defining how to construct models from reoccurring sub-components. Depending on the specific *instantiation*, these sub-components are duplicated to create the actual probabilistic model. We are interested in models that can be applied for reasoning about the relations between entities. Our motivating example will be reasoning about the structure of interaction networks (*e.g.*, social interaction networks or protein-protein interaction networks). We now define a class of relational models that will be convenient for reasoning about these domains. We define a language that is similar to ones previously defined [19], but also a bit different, to make our claims in the following section more clear.

As with most relational models in the literature we distinguish the *template-level* model that describe the types of objects and components of the model and how they can be applied, from the *instantiation-level* that describes a particular model which is an instantiation of the template to a specific set of entities.

To define a template-level model we first set up the different types of entities we reason about in the model. We distinguish *basic entity types* that describe atomic entities from *complex types* that describe composite entities.

**Definition 2.1:** Given a set $\mathcal{T}_{\text{basic}} = (T_1, \ldots, T_n)$ of *basic entity types* we define two kinds of **complex types**:

- If $T_1, \ldots, T_k$ are basic types, then $T_1 \times \cdots \times T_k$ denotes the type of *ordered tuples* of entities of these types. If $e_1, \ldots, e_k$ are entities of types $T_1, \ldots, T_k$, respectively, then $\langle e_1, \ldots, e_k \rangle$ is of type $T_1 \times \cdots \times T_k$.
- If $T$ is a basic type, then $T^k$ denotes the type of *unordered tuples* of entities of type $T$. If $e_1, \ldots, e_k$ are entities of type $T$, then $[e_1, \ldots, e_k]$ is of type $T^k$. When considering ordered tuples, permutations of the basic elements still refer to the same complex entity. Thus, if $e_1, e_2$ are of type $T$, then both $[e_1, e_2]$ and $[e_2, e_1]$ refer to the same complex entity of type $T^2$.

∎

For example, suppose we want to reason about undirected graphs. If we define a type $T_v$ for vertices then an undirected edge is of type $T_e \equiv T_v^2$ since an edge is a composite object that consists of two vertices. Note that we use unordered tuples since the edge does not have a direction. That is, both $[v_1, v_2]$ and $[v_2, v_1]$ refer to the same relationship between the two vertices. If we want to model directed edges, we need to reason about ordered tuples $T_e \equiv T_v \times T_v$. Now $\langle v_1, v_2 \rangle$ and $\langle v_2, v_1 \rangle$ refer to two distinct edges. We can also consider social networks, where vertices correspond to people. Now we might also add a type $T_l$ of physical locations. In order to reason about relationships between vertices (people) and locations we need to define pairs of type $T_p \equiv T_v \times T_l$. Note that tuples that relate between different types are by definition ordered.

Once we define the template-level set of types $\mathcal{T}$ over some set of basic types $\mathcal{T}_{\text{basic}}$, we can consider particular instantiations in terms of entities.

**Definition 2.2:** An *entity instantiation* $\mathcal{I}$ for $(\mathcal{T}_{\text{basic}}, \mathcal{T})$ consists of a set of *basic entities* $\mathcal{E}$ and a mapping $\sigma : \mathcal{E} \mapsto \mathcal{T}_{\text{basic}}$ that assigns a basic type to each basic entity. ∎

Based on an instantiation, we create all possible instantiations of each type in $\mathcal{T}$:

- if $T \in \mathcal{T}_{\text{basic}}$ then $\mathcal{I}(T) = \{e \in \mathcal{E} : \sigma(e) = T\}$

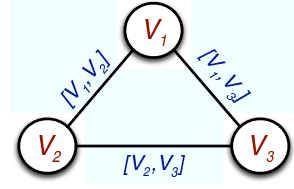

Figure 1: An instantiation of the graph scheme over a domain of three vertices.

- If $T = T_1 \times \cdots \times T_k$ then $\mathcal{I}(T) = \mathcal{I}(T_1) \times \cdots \times \mathcal{I}(T_k)$.
- If $T = T_1^k$ then $\mathcal{I}(T) = \{[e_1, \ldots, e_k] : e_1, \ldots e_k \in \mathcal{I}(T_1), e_1 \leq \cdots \leq e_k\}$ where $\leq$ is some (arbitrary) order over $\mathcal{I}(T)$ [1].

Once we define a set of basic entities, we assume that all possible complex entities of the given type are defined (see Figure 1 for an instantiation of the graph example).

The basic and complex entities define the structure of our domain of interest. Our goal, however, is to reason about the properties of these entities. We refer to these properties as *attributes*. Again, we start by the definition at the template level, and proceed to examine their application to a specific instantiation:

**Definition 2.3:** A **template attribute** $\mathsf{A}(T)$ defines a property of entities of type $T$. The set of values the attribute can take is denoted $Val(\mathsf{A}(T))$. ∎

A template attribute denotes a specific property we expect each object of the given type to have. In general, we can consider attributes of basic objects or attributes of complex objects. In our example, we can reason about the color of a vertex, by having an attribute $\mathsf{Color}(T_v)$. We can also create an attribute $\mathsf{Exist}(T_e)$ that denotes whether the edge between two vertices exists. We can consider other attributes such as the weight of an edge and so on. All these template attribute are defined at the level of the scheme and we will denote by $\mathcal{A}$ the set of template attributes in our model.

Given a concrete entity instance $\mathcal{I}$ we consider all the attributes of each instantiated type. We view the attributes of objects as random variables. Thus, each template attribute in $\mathcal{A}$ defines a set of random variables:

$$\mathcal{X}_\mathcal{I}(\mathsf{A}(T)) = \{X_\mathsf{A}(e) : e \in \mathcal{I}(T)\}$$

We define $\mathcal{X}_\mathcal{I} = \cup_{\mathsf{A}(T) \in \mathcal{A}} \mathcal{X}_\mathcal{I}(\mathsf{A}(T))$ to be the set of all random variables that are defined over the instantiation $\mathcal{I}$. For example, if we consider the attributes Color over vertices and Exist over unordered pairs of vertices, and suppose that $\mathcal{E} = \{v_1, v_2, v_3\}$ are all of type $T_v$, then we have three random variables in $\mathcal{X}(\mathsf{Color}(T_v))$ which are $X_{\mathsf{Color}}(v_1), X_{\mathsf{Color}}(v_2), X_{\mathsf{Color}}(v_3)$, and four random variables in $\mathcal{X}(\mathsf{Exist}(T_e))$ which are $X_{\mathsf{Exist}}([v_1, v_2]), X_{\mathsf{Exist}}([v_1, v_3])$, and so on.

---

[1] For example, considering undirected edges again, we think of $[v_1, v_2]$ and $[v_2, v_1]$ as two different names of the same entity. Our definition ensures that only one of these two objects is in the set of entities and we view the other as an alternative reference to the same entity.

Given a set of types, their attributes and an instantiation, we defined a universe of discourse, which is the set $\mathcal{X}_\mathcal{I}$ of random variables. An *attribute instantiation* $\omega$ (or just instantiation) is an assignment of values to all random variables in $\mathcal{X}_\mathcal{I}$. We use both $\omega(X_\mathsf{A}(e))$ and $x_\mathsf{A}(e)$ to refer to the assigned value to the attribute $\mathsf{A}$ of the entity $e$.

We now turn to the final component of our relational model. To define a log-linear model over the random variables $\mathcal{X}_\mathcal{I}$, we need to introduce *features* that capture preferences for specific combinations of values to small groups of related random variables. In our graph example, we can introduce a univariate feature for edges that describes the prior potential for the existence of an edge in the graph. A more complex feature can describe preferences over triplets of interactions (*e.g.*, prefer triangles over open chains).

We start by defining template level features as a recipe that will be assigned to a large number of specific sets of random variables in the instantiated model. Intuitively, a template feature defines a function that can be applied to a set of attributes of related entities. To do so, we need to provide a mechanism to capture sets of entity attributes with particular relationships. For example, to put a feature over triangle-like edges, we want a feature over the variables $X_{\mathsf{Exist}}([v_1, v_2])$, $X_{\mathsf{Exist}}([v_1, v_3])$, and $X_{\mathsf{Exist}}([v_2, v_3])$ for every choice of three vertices $v_1, v_2$, and $v_3$. The actual definition, thus involves entities that we quantify over (e.g., $v_1, v_2$, and $v_3$), the complex entities over these arguments we examine (e.g., $[v_1, v_2], [v_1, v_3]$, and $[v_2, v_3]$), the attributes of these entities, and the actual feature.

**Definition 2.4: Template Feature** A *template feature* $\mathcal{F}$ is defined by four components:

- A tuple of *arguments* $\langle \xi_1, \ldots, \xi_k \rangle$ with a corresponding list of *type signature* $\langle T_1^q, \ldots, T_k^q \rangle$, such that $\xi_i$ denotes an entity of basic type $T_i^q$.

- A list of formal entities $\varepsilon_1, \ldots, \varepsilon_j$, with corresponding types $T_1^f, \ldots, T_j^f$ such that each formal entity $\varepsilon$ is either one of the arguments, or a complex entity constructed from the arguments. (For technical reasons, we require that formal entities refer to each argument at most once.)

- A list of attributes $\mathsf{A}_1(T_1^f), \ldots, \mathsf{A}_j(T_j^f)$.

- A function $f : Val(\mathsf{A}_1(T_1^f)) \times \cdots \times Val(\mathsf{A}_j(T_j^f)) \mapsto \mathbb{R}$.

For example, Table 1 shows such a formalization for a graph model with two such template level features.

| | Arguments | Formal entities | Attr. | Function |
|---|---|---|---|---|
| $\mathcal{F}_e$ | $\langle \xi_1, \xi_2 \rangle$ $\langle T_v, T_v \rangle$ | $[\xi_1, \xi_2]$ $T_e$ | Exist | $f_\delta(z) = \mathbf{1}\{z = 1\}$ |
| $\mathcal{F}_t$ | $\langle \xi_1, \xi_2, \xi_3 \rangle$ $\langle T_v, T_v, T_v \rangle$ | $[\xi_1, \xi_2]$ $[\xi_1, \xi_3]$ $[\xi_2, \xi_3]$ $T_e, T_e, T_e$ | Exist Exist Exist | $f_3(z_1, z_2, z_3) =$ $\mathbf{1}\{(z_1 = 1) \wedge$ $(z_2 = 1) \wedge$ $(z_3 = 1)\ \}$ |

Table 1: Example of two template-level features for a graph model. The first is a feature over single edges, and the second is one over triplets of coincident edges (triangles).

We view a template-level feature as a recipe for generating multiple instance-level features by applying different *bindings* of objects to the arguments. For example, in our three vertices instantiation, we could create instances of the feature $\mathcal{F}_e$ such as $f_\delta(X_{\mathsf{Exist}}([v_1, v_2]))$ and $f_\delta(X_{\mathsf{Exist}}([v_1, v_3]))$. We now formally define this process.

**Definition 2.5:** Let $\mathcal{F}$ be a template feature with components as in Definition 2.4, and let $\mathcal{I}$ be an entity instantiation. A *binding* of $\mathcal{F}$ is an ordered tuple of $k$ entities $\beta = \langle e_1, \ldots, e_k \rangle$ such that $e_i \in \mathcal{I}(T_i^q)$. A binding is *legal* if each entity in the binding is unique. We define

$$\mathrm{Bindings}(\mathcal{F}) = \{\beta \in \mathcal{I}(T_1^q) \times \cdots \times \mathcal{I}(T_k^q) : \beta \text{ is legal for } \mathcal{F}\}$$

Given a binding $\beta = \langle e_1, \ldots, e_k \rangle \in \mathrm{Bindings}(\mathcal{F})$, we define the entity $\varepsilon_i|_\beta$ to be the entity corresponding to $\varepsilon_i$ when we assign $e_i$ to the argument $\xi_i$. Finally, we define the *ground feature* $\mathcal{F}|_\beta$ to be the function over $\omega$:

$$\mathcal{F}|_\beta(\omega) = f\left(\omega(X_{\mathsf{A}_1}(\varepsilon_1|_\beta)), \ldots, \omega(X_{\mathsf{A}_j}(\varepsilon_j|_\beta))\right)$$

For example, consider the binding $\langle v_1, v_2, v_3 \rangle$ for $\mathcal{F}_t$ of Table 1. This binding is legal since all three entities are of the proper type and are different from each other. This binding defines the ground feature

$$\mathcal{F}_t|_{\langle v_1, v_2, v_3 \rangle}(\omega) =$$
$$f_3(x_{\mathsf{Exist}}([v_1, v_2]), x_{\mathsf{Exist}}([v_1, v_3]), x_{\mathsf{Exist}}([v_2, v_3]))$$

That is, $\mathcal{F}_t|_{\langle v_1, v_2, v_3 \rangle}(\omega) = 1$ iff there is a triangle of edges between the vertices $v_1, v_2$, and $v_3$. Note that each binding defines a ground feature. However, depending on the choice of feature function, some of these ground features might be equivalent. In our last example, the binding $\langle v_1, v_3, v_2 \rangle$ creates the same feature. While this creates a redundancy, it does not impact the usefulness of the language. We now have all the components in place.

**Definition 2.6:** A *Relational MRF scheme* $\mathcal{S}$ is defined by a set of types $\mathcal{T}$, their attributes $\mathcal{A}$ and a set of template features $\mathscr{F} = \{\mathcal{F}_1, \ldots, \mathcal{F}_k\}$. A *model* is a scheme combined with a vector of *parameters* $\theta = \langle \theta_i, \ldots, \theta_k \rangle \in \mathbb{R}^k$. Given an entity instantiation $\mathcal{I}$ a scheme uniquely defines the universe of discourse $\mathcal{X}_\mathcal{I}$. Given all this together we can define the joint distribution of a full assignment $\omega$ as:

$$P(\omega : \mathcal{S}, \mathcal{I}, \theta) = \frac{1}{Z(\theta, \mathcal{I})} \exp \sum_{i=1}^{k} \theta_i \mathcal{F}_i(\omega) \quad (1)$$

where (with slight abuse of notation)

$$\mathcal{F}_i(\omega) = \sum_{\beta \in \text{Bindings}(\mathcal{F}_i)} \mathcal{F}_i|_\beta(\omega)$$

is the total weight of all grounding of the feature $\mathcal{F}_i$, and $Z$ is the normalizing constant. ∎

This definition of a joint distribution is similar to standard log-linear models, except that all groundings of a template feature share the same parameter [4].

## 3  Compact Approximate Inference

One broad class of approximate inference procedure are *variational methods* [12]. Roughly speaking, in such methods we approximate the joint distribution by introducing additional *variational* parameters. Depending on the particular method, these additional parameters can be thought of as capturing approximation of marginal beliefs about selected subsets of variables. Although the general idea we present here can be applied to almost all variational methods, for concreteness and simplicity we focus here on *loopy belief propagation* [16, 23] which is one of the most common approaches in the field.

To describe loopy belief propagation we consider the data structure of a *factor graph* [14]. A factor graph is a bipartite graph that consists of two layers. In the first layer, we have for each random variable in the domain a *variable node* $X$. In the second layer we have *factor nodes*. Each factor node $\omega$ is associated with a set $\mathbf{C}_\omega$ of random variables and a feature $\pi_\omega$. If $X \in \mathbf{C}_\omega$, then we connect the variable node $X$ to the factor node $\omega$. Graphically we draw variable nodes as circles and factor nodes as squares (see Figure 2(a)).

A factor graph is *faithful* to a log-linear model if each feature is assigned to a node whose scope contains the scope of the feature. Adding these features multiplied by their parameters defines for each potential node $\omega$ a potential function $\pi_\omega[\mathbf{c}_\omega]$ that assigns a real value for each value of $\mathbf{C}_\omega$. There is usually a lot of flexibility in defining the set of potential nodes. For simplicity, we focus now on factor graphs where we have a factor node for each ground feature.

For example, let us consider a model over a graph where we also depict the colors of the vertices. We create for each vertex $v_i$ a variable node $X_{\text{Color}}(v_i)$ and for each pair of vertices $[v_i, v_j]$ a variable node $X_{\text{Exist}}([v_i, v_j])$. We consider two template features - the triangle feature we described earlier, and a co-colorization feature that describes a preference of two vertices that are connected by an edge to have the same color. To instantiate the triangle feature, we go over all directed tuples of three vertices $\beta = \langle v_i, v_j, v_k \rangle \in \text{Bindings}(\mathcal{F}_t)$ and define $\omega_\beta$ with scope $\mathbf{C}_\beta = \{X_{\text{Exist}}([v_i, v_j]), X_{\text{Exist}}([v_i, v_k]), X_{\text{Exist}}([v_j, v_k])\}$. See Figure 2(a) to see such a factor graph for an instantiation of 4 vertices. This factor graph is faithful since each ground feature is assigned to a dedicated feature node.

Loopy belief propagation over a factor graph is defined as repeatedly updating messages of the following form:

$$m_{X \to \omega}(x) \leftarrow \prod_{\omega': X \in \mathbf{C}_{\omega'}, \omega' \neq \omega} m_{\omega' \to X}(x)$$

$$m_{\omega \to X}(x) \leftarrow \sum_{\mathbf{c}_\omega \langle X \rangle = x} \left( e^{\pi_\omega[\mathbf{c}_\omega]} \prod_{X \neq X' \in \mathbf{C}_\omega} m_{X' \to \omega}(x') \right)$$

where $\mathbf{c}_\omega \langle X \rangle$ is the value of $X$ in the assignment of values $\mathbf{c}_\omega$ to $\mathbf{C}_\omega$. When these messages converge, we can define belief about variables as

$$b_\omega(\mathbf{c}_\omega) \propto e^{\pi_\omega[x]} \prod_{X' \in \mathbf{C}} m_{X \to \omega}(\mathbf{c}_\omega \langle X' \rangle)$$

where the beliefs over $\mathbf{C}_\omega$ are normalized to sum to 1. These beliefs are the approximation of the marginal probability over the variables in $\mathbf{C}_\omega$ [23].

Unfortunately, trying to reason about a network over 1000 vertices with the features we described earlier, will produce $\binom{1000}{2}$ variable nodes (one for each edge), $2 \cdot \binom{1000}{2}$ edge feature nodes and $3 \cdot \binom{1000}{3}$ triplet feature nodes[2]. Building such a graph and performing loopy belief propagation with it is a time consuming task. However, our main insight is that we can exploit some special properties of this model for much efficient representation and inference. The basic observation is that the factor graphs for the class of models we defined satisfy basic symmetry properties.

Specifically, consider the structure of the factor graph we described earlier. An instantiation of graph vertices defines both the list of random variables and of features that will be created. Each feature node represents a ground feature that originates from a legal binding to a template feature. The groundings for an edge feature and for an edge random variable span two vertices, while the grounding of triplet feature covers three vertices. Since we are considering all legal bindings (*i.e.*, all 2-mers and 3-mers of vertices) while spanning the factor graph, each edge variable node will be included in the scope of 2 edge feature nodes and $(n-2) \cdot 3$ triplet feature nodes. More importantly, since all the edge variables have the same "local neighborhood", they will also compute the same messages during belief propagation over and over again. We now formalize this idea and show we can use it to enable efficient representation and inference.

---

[2]Since we defined the template feature using ordered tuples and our edges are defined using unordered tuples, we will have two features over each edge and three features over each triplet.

**Definition 3.1:** We say that two nodes in the factor graph have the same **type** if they were instantiated from the same template (either template attribute or template feature). ∎

Given this definition, we can present our main claim formally:

**Theorem 3.2:** In every stage $t$ of synchronous belief propagation that is initiated with uniform messages, if $v_i, v_k$ are from the same type and also $v_j, v_l$ are from the same type then $m^t_{v_i \to v_j}(x) = m^t_{v_k \to v_l}(x)$.

We start by proving the local properties of symmetry of the model:

**Lemma 3.3:** In a model created according to Definition 2.6, if two nodes in the factor graph have the same type, then they have the same **local neighborhood**. That is, they have the same number of neighbors of each type.

The proof of Theorem 3.2 is a direct consequence of Lemma 3.3 by induction over the stage of the belief propagation. We now turn to prove Lemma 3.3:

**Proof:** If $v_i$ and $v_j$ are feature nodes, then since they are of the same type, they are instantiations of the same template feature. From Definition 2.4 and Definition 2.5 we can see that this means that they are defined over variables from the same type. Since each feature is connected only to the variables in its scope, this proves our claim. However, if $v_i$ and $v_j$ are variable nodes, it suffices to show that they take part in the same kind of features, and in the same number of features of each such kind. Note that Definition 2.6 shows that we use all legal binding for each feature. For simplicity, we will assume that $v_i$ is instantiated from the attribute of some basic type $T$ (the proof in case it is a complex type is similar). We need to compute how many ground features contain $v_i$ in their scope, and do not contain $v_j$. From Definition 2.5 we can see that all the legal bindings that include $v_i$ and do not include $v_j$ are legal also if we replace $v_i$ with $v_j$. ∎

After showing that many calculations are done over and over again, we now show how we can use a more efficient representation to enable much faster inference.

**Definition 3.4:** A **template factor graph** over a template log-linear model is a bi-partite graph, with one level corresponding to attributes and the other corresponding to template features. Each template attribute T that corresponds to a formal entity in some template feature $\mathcal{F}$ is mapped to a *template attribute node* on one side of the graph. And each template feature is mapped to a *template feature node* on the other side of the graph. Each template attribute node is connected with an edge to all the template feature nodes that contain this feature in their scope. A feature node needs to distinguish between its neighbors, since each message refers to a message about different variable. Hence, in the template factor graph we term an association to a variable inside a template feature node *port* . If a factor contains more than one variable of the same type, the corresponding edge splits to the corresponding ports when arriving to the factor node. In addition, each ground variable node takes part in many features that were instantiated by the same template feature with different bindings. Hence, each edge from a template feature node to a template attribute node in the template factor graph is assigned with a number indicating the number of repetitions it has in the full factor graph. ∎

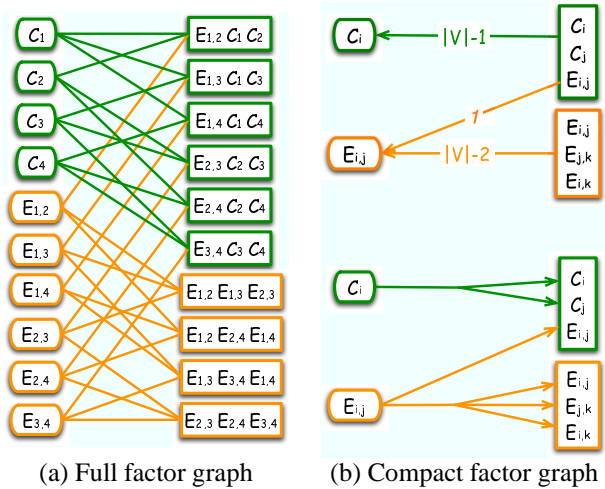

(a) Full factor graph　　(b) Compact factor graph

Figure 2: Shown are the full (a) and compact (b) factor graphs modeling a colored graph. We have basic types for colors and vertices, and a complex type for edges. We consider two template features - the triangle feature and a co-colorization feature. For clarity, $X_{\text{Exist}}([v_i, v_j])$ is shown as $E_{i,j}$ and $X_{\text{Color}}(v_i)$ is shown as $C_i$. Orange edges show the edges connected to edge variables and green edges are connected to color variables. $|V|$ shows the number of vertices in the graph.

Figure 2(b) shows such a template factor graph for our running example.

Running loopy belief propagation on this template factor graph is straightforward. The algorithm is similar to the standard belief propagation only that when an edge in the template-graph represents many edges in the instance-level factor graph, we interpret this by multiplying the appropriate message the appropriate number of times. Since Theorem 3.2 shows that at all stages in the standard synchronous belief propagation the messages between nodes of the same type are similar, running belief propagation on the template factor graph is equivalent to running synchronous belief propagation on the full factor graph. However, we reduced the cost of representation and inference from being proportional to the size of the instantiated model, to be propor-tional to the size of the domain. Specifically, this representation does not depend on the size of the instantiations and can deal with a huge number of variables.

## 4 Evaluation

### 4.1 Inference

We start by evaluating our method in inference tasks. We build a model representing a graph using the univariate and triangle features described in the previous section and perform inference with various parameter combinations. In the first step we consider instantiations of small graphs where we can also perform exact inference. We compared exact inference, MCMC (Gibbs sampling) [8], standard asynchronous belief propagation [23], and compact belief propagation on the template-level model. A simple way to compare inference results is by examining the marginal beliefs. Such a comparison is possible since in all methods the computed marginal probabilities for all edge variables were equal. Hence, Figure 3 shows a comparison of the marginal distributions over edge variables for different parameter settings and different inference methods. We observe that in small graphs the marginal beliefs are very similar for all inference methods. To quantify the similarity we calculate the relative deviation from the true marginal. We find that on average MCMC deviates by $0.0118$ from the true marginal (stdev: $0.0159$), while both belief propagation methods deviate on average by $0.0143$ (stdev: $0.0817$) and are virtually indistinguishable. However, in the graph over 7 vertices we notice that exact inference and MCMC are slightly different from the two belief propagation methods in the case where the univariate parameter is small and the triplet parameter is large (lower right corner).

An alternative measurement of inference quality is the estimate of the partition function. This is especially important for learning applications, as this quantity serves to compute the likelihood function. When performing loopy belief propagation, we can approximate the log-partition function using the Bethe approximation [23]. As seen in Figure 4, the estimate of the log partition function by belief propagation closely tracks the exact solution. Moreover, as in the marginal belief test, the two variants of belief propagation are almost indistinguishable. It is important to stress that running times are substantially different between the methods. For example, using exact inference with the 7 vertices graph (*i.e.*, one pixel in the matrices shown in Figure 3) takes $80$ seconds on a $2.4$ GHz Dual Core AMD based machine. Approximating the marginal probability using MCMC takes $0.3$ seconds, standard BP takes $12$ seconds, and compact BP takes $0.07$ seconds.

On larger graphs, where exact inference and standard belief propagation are infeasible, we compare only the compact belief propagation and MCMC (see Figure 5). While there are some differences in marginal beliefs, we see again that in general there is good agreement between the two inference procedures. As the graph becomes larger the gain in run-time increases. Since the mixing time of MCMC should depend on the size of the graph (if accuracy is to be conserved), running MCMC inference on a 100-node graph takes $5$ minutes. As expected, compact BP still runs for only $0.07$ seconds since it depends on the size of the scheme which remains the same. For protein-protein interaction networks over hundreds of vertices (see below), all inference methods become infeasible except for compact belief propagation.

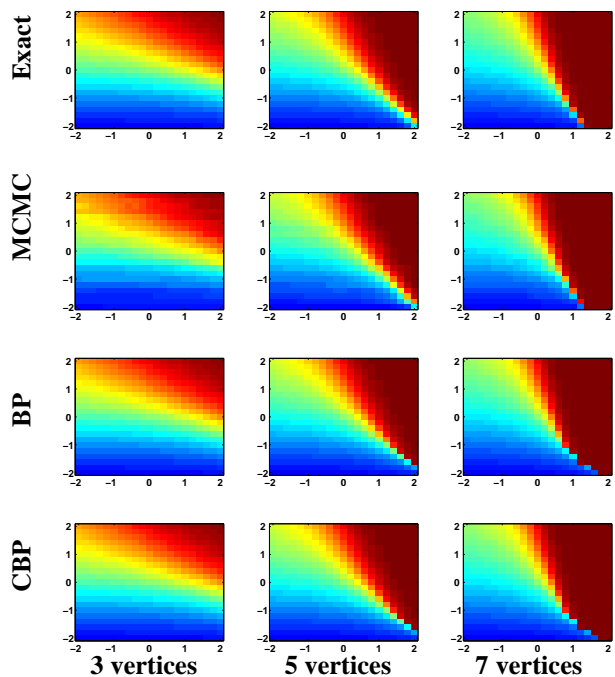

Figure 3: Comparison of inference methods via marginal beliefs. Each panel visualizes the the probability of an interaction when we vary two parameters: the univariate potential for interaction ($y$-axis) and the the potential over closed triplet ($x$-axis). The color indicates probability where blue means probability closer to $0$ and red means probability closer to $1$. The first row of panels shows exact computation, the second MCMC, the third standard asynchronous belief propagation, and the fourth our compact belief propagation.

### 4.2 Parameter estimation

Consider the task of learning the parameters $\Theta = \langle \theta_1 \ldots \theta_k \rangle$ for each template feature. To learn such parameters from real-life data we can use the *Maximum Likelihood* (ML) estimation [4]. In this method we look for the parameters that best explain the data in the sense that they find $\mathrm{argmax}_{\theta \in \Theta} p(\boldsymbol{D}|\theta)$. Since there is no closed form for finding the maximum likelihood parameters of a log-linear model, a common approach is to resort to greedy search methods such as gradient ascent. In such approaches an efficient calculation of the derivative is needed. The partial

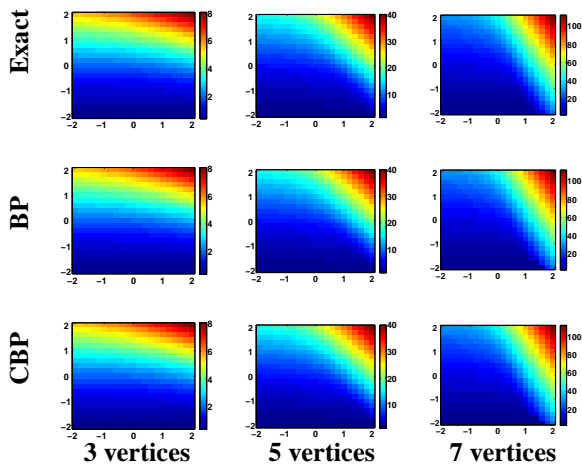

3 vertices     5 vertices     7 vertices derivative of the log likelihood $\ell(\boldsymbol{D})$ for a parameter $\theta_j$ that corresponds to a template feature $\mathcal{F}_j$ can be described as:

$$\frac{\partial \ell(\boldsymbol{D})}{\partial \theta_j} = \hat{\boldsymbol{E}}\left[\mathcal{F}_j\right] - \boldsymbol{E}_\theta\left[\mathcal{F}_j\right] \quad (2)$$

Figure 4: Comparison of inference methods for computing the log-partition function. Each panel visualizes the log-partition function (or its approximation) for different parameter setting (as in Figure 3). In the belief propagation methods, the log-partition function is approximated using the Bethe free energy approximation. On the first row is the exact computation, the second row shows standard asynchronous belief propagation and the third row shows our compact belief propagation.

Where $\hat{\boldsymbol{E}}\left[\mathcal{F}_j\right]$ is the number of times we actually see the feature $j$ in $\boldsymbol{D}$, and

$$\boldsymbol{E}\left[\mathcal{F}_j\right] = \sum_{\beta \in \text{Bindings}(\mathcal{F}_j)} \boldsymbol{E}\left[\mathcal{F}_j|\beta\right]$$

is the sum of times we expect to see each grounding of the feature $j$ according to $\Theta$ (see [4]). The first term is relatively easy to compute in cases where we learn from fully observed instances, since it is simply the count of each feature in $\boldsymbol{D}$. And the second term can be approximated efficiently by our inference algorithm.

To evaluate this learning procedure we start by generating samples from a model using a Gibbs sampler [8]. We then use these samples to estimate the original parameters using exact and approximate inference. In this synthetic context, we model a graph over seven vertices using only triplet ($\mathcal{F}_t$) and open chain ($\mathcal{F}_c$) features and try to recover the parameter of these features. As can be seen in Figure 6, using both approximate and exact inference retrieved parameter values that are close to these we used to generate the data. However, we can see that since the approximate and exact likelihoods create a different scenery, the trace of the exact search is much shorter, and retrieves better parameters.

We now proceed to learning a real-life model over interactions between proteins. We build on a model described

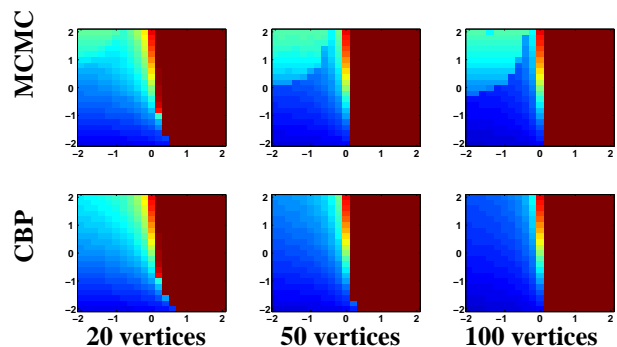

20 vertices     50 vertices     100 vertices

Figure 5: Comparison of approximate inference methods on larger graph instances. As before, we show the probability of an interaction as a function of parameter settings. On the first row is MCMC and the second row shows our compact belief propagation.

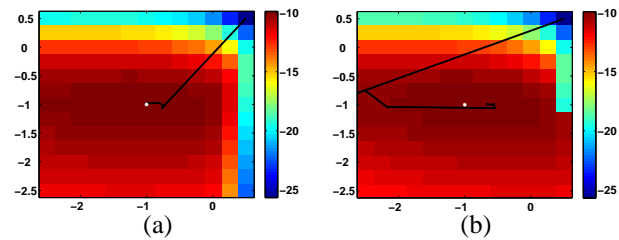

(a)                 (b)

Figure 6: Learning trace of the parameters using exact (a) and approximate (b) inference on a 7 vertex graph. In both panels values of $\theta_{111}$ are shown on the $x$-axis while values of $\theta_{011}$ are shown on the $y$-axis. The dark line shows the advancement of the conjugate gradient learning procedure, and the bright asterix in the middle shows the original parameters used for generating the samples. Color scale shows the exact and approximate log-likelihood respectively

in [11] for protein-protein interactions. This model is analogous to our running example, where the vertices of the graph are proteins and the edges are interactions. We define the basic type $T_p$ for proteins and the complex type $T_i = [T_p, T_p]$ for interactions between proteins. As with edges, we consider the template attribute $\mathsf{X}_e(T_i)$ that equals one if the two proteins interact and zero otherwise. We reason about an instantiation for a set of 813 proteins related to DNA transcription and repair [2]. We collected statistics over interactions between these proteins from various experiments [1, 7, 13, 15].

We adopt an incremental approach considering only the simplest template feature at the beginning and adding more complex features later on (this approach is somewhat similar to Della Pietra *et al.* [4]). We start by learning a model with only univariate features over interactions. As expected, the parameters we learn reflect the probability of an interaction in the data. We can now consider more complex features to the model by fixing the univariate parameter and adding various features. We start by adding two features, $\mathcal{F}_t$ and $\mathcal{F}_c$ that describe the closed triangle of interactions and open chain of interactions respectively.

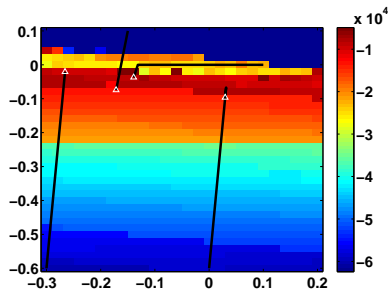

Figure 7: Exploration of the approximate log-likelihood landscape. In this example, the univariate parameter is fixed, the weights of two features over three interactions, triangle and chains, are varied. The $x$-axis shows the triangle parameter ($\theta_{111}$) and the $y$-axis shows the chain parameter ($\theta_{011}$). The dark lines show traces of conjugate gradient runs initiated from arbitrary starting points. The bright triangles mark the final parameter values returned by the algorithm.

Using our efficient inference approximation we can reevaluate the likelihood and its derivative for many parameter values and thereby gain an unprecedented view of the likelihood landscape of the model. For example, Figure 7 shows the log-likelihood calculated for a grid of parameter values and traces of a conjugate gradient learning procedure initialized from different starting points. We find that this view of the likelihood function is highly informative as it shows the influence of different parameter values on the model behavior. Specifically, the results show that the likelihood sensitivity to each parameter is quite different. This can be seen as a horizontal ridge in the upper part of the region, meaning that changes in $\theta_{111}$ have smaller effect on likelihood value than changes in $\theta_{011}$. This behavior might reflect the fact that there are 3-times more occurrences of open chains than occurrences of closed triangles in the graph. Furthermore, our unique view of the likelihood landscape, and especially the horizontal ridge we see, illustrate that there is a strong relation between the parameters. As each of the gradient ascent runs converge to a different local maxima, we can use the landscape to determine whether this a consequence of rough landscape of the approximate likelihood or is due to redundancies in the parametrization that result in an equi-probable region.

We repeated the same exploration technique for other features such as colocalization of proteins [11], star-2 and star-3 [10], and quadruplets of interactions (results not shown). We find that the overall gain in terms of likelihood is smaller than in the case of triplet features. Again, we find that whenever one of the features is more abundant in the network, its influence on the approximate marginal beliefs and likelihood is much larger. In such cases the interesting region - where likelihood is high - narrows to a small range of parameter values of the abundant feature.

## 5 Discussion

We have shown how we exploit symmetry in relational MRFs to perform approximate inference at the template-level. This results in an extremely efficient approximate inference procedure. We have shown that this procedure is equivalent to synchronous belief propagation in the ground model. We have also empirically shown that on small graphs our inference algorithm approximates the true marginal probability very well. Furthermore, other approximation methods, such as MCMC and asynchronous BP yield inference results that are similar to ours. Note that other works show that synchronous and asynchronous belief propagation are not always equivalent [5].

Other works attempted to exploit relational structure for more efficient inference. For example, Pfeffer *et al.* [17] used the relational structure to cache repeated computations of intermediate terms that are identical in different instances of the same template. Several recent works [3, 18] derive rules as to when variable elimination can be performed at the template level rather than the instance level, which saves duplicate computations at the instance levels. These methods focus on speeding exact inference, and are relevant in models where the intermediate calculations of exact inference have tractable representations. These approaches cannot be applied to models, such as the ones we consider, where the tree-width is large, and thus intermediate results of variable elimination are exponential. In contrast, our method focuses on template level inference for approximate inference in such intractable models.

We stress that the main ideas developed here can be applied in other variational methods such as generalized belief propagation or structured mean field. Furthermore, it is clear that the class of relational models we defined is not the only one that has symmetry properties that can be exploited by our procedure. In fact, all the relational models that obey Lemma 3.3 can be run in template level. For example, it can be shown that a square wrap-around grid also obeys such symmetry.

The key limitation of our procedure is that it relies on the lack of evidence. Once we introduce evidence the symmetry is disrupted and our method does not apply. While this seems to be a serious limitation, we note that inference without evidence is the main computational step in learning such models from data. We showed how this procedure enables us to deal with learning problems in large relational models that were otherwise infeasible. Though the search space proves to be very difficult [10], our method allows us to perform many iterations of parameter estimation in different settings and thereby get a good overview of the likelihood landscape. This brings us one step closer towards successful modeling of networks using relational probabilistic models.

**Acknowledgements**

We thank Chen Yanover, Tal El-Hay, Gal Elidan, and the anonymous reviewrs for helpful remarks on previous versions of this manuscript. Part of this research was supported by a grant from the United States-Israel Binational Science Foundation (BSF). Ariel Jaimovich is supported by the Eshkol fellowship from the Israeli Ministry of Science.

## References

[1] S. R. Collins, *et al*. Towards a comprehensive atlas of the physical interactome of Saccharomyces cerevisiae. *Mol Cell Proteomics*, 2007.

[2] S. R. Collins, *et al*. Functional dissection of protein complexes involved in yeast chromosome biology using a genetic interaction map. *Nature*, 2007.

[3] R. de Salvo Braz, E. Amir, and D. Roth. Lifted first-order probabilistic inference. In *IJCAI* 2005.

[4] S. Della Pietra, V. Della Pietra, and J. Lafferty. Inducing features of random fields. *IEEE Trans. on Pattern Analysis and Machine Intelligence*, 19(4):380–393, 1997.

[5] G. Elidan, I. McGraw, and D. Koller. Residual belief propagation: Informed scheduling for asynchronous message passing. In *UAI* 2006.

[6] N. Friedman, L. Getoor, D. Koller, and A. Pfeffer. Learning probabilistic relational models. In *IJCAI* 1999.

[7] A. C. Gavin, *et al*. Proteome survey reveals modularity of the yeast cell machinery. *Nature*, 440(7084):631–636, 2006.

[8] S. Geman and D. Geman. Stochastic relaxation, gibbs distributions, and the bayesian restoration of images. *IEEE Trans. on Pattern Analysis and Machine Intelligence*, pages 721–741, 1984.

[9] L. Getoor, N. Friedman, D. Koller, and B. Taskar. Learning probabilistic models of relational structure. In *ICML* 2001.

[10] M. S. Handcock. Assessing degeneracy in statistical models of social networks. Technical Report 39, University of Washington, 2003.

[11] A. Jaimovich, G. Elidan, H. Margalit, and N. Friedman. Towards an integrated protein-protein interaction network: a relational Markov network approach. *J. Comut. Biol.*, 13:145–164, 2006.

[12] M. I. Jordan, Z. Ghahramani, T. Jaakkola, and L. K. Saul. An introduction to variational approximations methods for graphical models. In *Learning in Graphical Models*, 1998.

[13] N. J. Krogan, *et al*. Global landscape of protein complexes in the yeast Saccharomyces cerevisiae. *Nature*, 440(7084):637–643, 2006.

[14] F. R. Kschischang, B. J. Frey, and H. A. Loeliger. Factor graphs and the sum-product algorithm. *IEEE Transactions on Information Theory*, 47(2), 2001.

[15] HW Mewes, J Hani, F Pfeiffer, and D Frishman. MIPS: a database for genomes and protein sequences. *Nucleic Acids Research*, 26:33–37, 1998.

[16] K. Murphy and Y. Weiss. Loopy belief propagation for approximate inference: An empirical study. In *UAI* 1999.

[17] A. Pfeffer, D. Koller, B. Milch, and K. Takusagawa. SPOOK: A system for probabilistic object-oriented knowledge representation. In *UAI* 1999.

[18] D. Poole. First-order probabilistic inference. In *IJCAI* 2003.

[19] M. Richardson and P. Domingos P. Markov logic networks. *Machine Learning*, 62:107–136, 2006.

[20] E. Segal, M. Shapira, A. Regev, D. Pe'er, D. Botstein, D. Koller, and N. Friedman. Module networks: identifying regulatory modules and their condition-specific regulators from gene expression data. *Nat Genet*, 34(2):166–176, 2003.

[21] B. Taskar, A. Pieter Abbeel, and D. Koller. Discriminative probabilistic models for relational data. In *UAI* 2002.

[22] B. Taskar, M. F. Wong, P. Abbeel, and D. Koller. Link prediction in relational data. In *NIPS* 2004.

[23] J. Yedidia, W. Freeman, and Y. Weiss. Constructing free energy approximations and generalized belief propagation algorithms. Technical Report TR-2002-35, Mitsubishi Electric Research Laboratories, 2002.